# A SLEEP MONITORING SYSTEM BASED ON AUDIO, VIDEO AND DEPTH INFORMATION

[1] *Lyn Chao-ling Chen (陳昭伶)*, [2] *Kuan-Wen Chen (陳冠文)*, [3] *Yi-Ping Hung (洪一平)*

[1,3] Graduate Institute of Networking and Multimedia,
National Taiwan University, Taipei

[2] Intel-NTU Connected Context Computing Center, Taipei
E-mail: lynchen@ntu.edu.tw

## ABSTRACT

For quantitative evaluation of sleep disturbances, a non-invasive monitoring system is developed by introducing an event-based method. We observe sleeping in home context and classify the sleep disturbances into three types of events: motion events, light-on/off events and noise events. A device with an infrared depth sensor, a RGB camera, and a four-microphone array is used in sleep monitoring in an environment with barely light sources. One background model is established in depth signals for measuring magnitude of movements. Because depth signals can't observe lighting changes, another background model is established in color images for measuring magnitude of lighting effects. An event detection algorithm is used to detect occurrences of events from the processed data of the three types of sensors. The system was tested in sleep condition and the experiment result validates the system reliability.

***Keywords*** *Image Sequence Analysis; Event Detection; Non-invasive Sleep Monitoring*

## 1. INTRODUCTION

Sleep disturbances in a sleep environment affect sleep quality, likewise a bad mood causes trouble sleeping, and that both appear in motion behavior during sleep. We classify the sleep disturbances in home context into three types of events: motion events, light-on/off events and noise events. For detecting these events, the detection interface of our sleep monitoring system shows the data streams from three types of sensors, and the events are detected and recorded that users can navigate them via the browsing interface (Figure 1).

### 1.1. Sleep disturbances in home context

During sleeping, people are unaware of disturbances in the sleep environments, and the consequences are related to their sleep qualities. They feel the need to be assured their sleeping in good sleep conditions. Ideal sleep is consisted of falling asleep quickly, overnight sleeping, and refreshed experience of waking up [1].

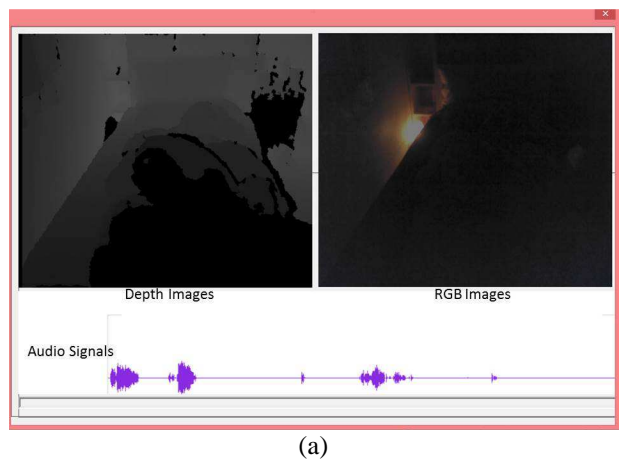

(a)

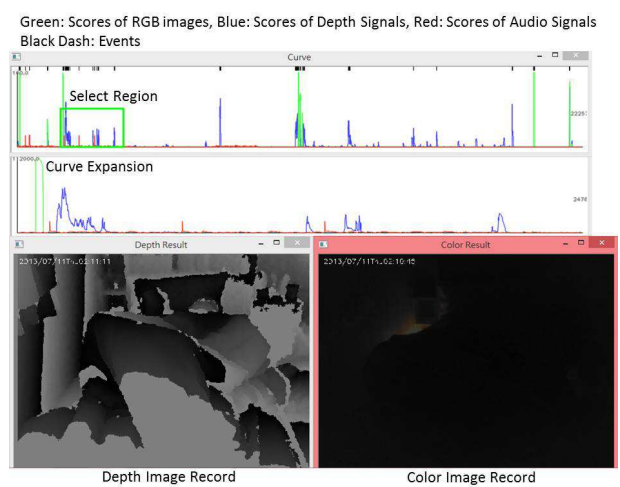

(b)

Fig. 1: Sleep Monitoring System: (a) Detection Interface, (b) Browsing Interface

Sometimes, it is interrupted by sleep disturbances that results in the circumstances of falling asleep difficulty, waking up too early, and staying asleep [2]. In an optimal sleep environment, darkness and tranquility are basic conditions to most of people [3]. Sources of lighting and noises from outside traffic noises, or talking on a street affect sleeping. Usually, the more serious problem is caused near by a partner who sleeps in the same room with different sleep schedule, like reading, or typing on a computer [4]. Specially, a snoring partner seriously disturbs the sleep of other people [5]. Considering other conditions, a comfortable temperature for sleep differs individually [4]. Besides, a bed with suitable firmness helps sleep, but it also differs by different shapes and sizes of people [6]. In this study, we consider lighting and sound conditions in a sleep environment.

### 1.2. Motion behavior during sleeping

Observing sleeping, people with sleep disorder problems clearly have restless sleep with many body movements. Normally, people shift their body positions within 40 to 60 per night, and they have many twitches in the sleep phase of rapid eye movement (REM) [7]. Occasionally, they have periodic limb movements during sleeping. Besides, people with mood disorders, or busy mind with various thoughts, ideas, and worries often have trouble sleeping [2] [8]. Moreover, unaware twitches also happen in a bad dream.

For quantitative evaluation of sleep disturbances, we develop a sleep monitoring system by an event-based method to detect motion events, light-on/off events and noise events.

## 2. RELATED WORKS

Sleep quality can be measured subjectively and objectively. Questionnaires, interviews and observations are the most common subjective measurement techniques. The objective measurement technology uses monitoring equipment to translate sleeping into a quantitative evaluation that we focus it on this study.

### 2.1. Bio-information for sleep quality measurement

Stanford University Sleep Disorders Clinic, the first medical clinic with equipment for specializing in sleep disorders was founded in the early 1970s [9]. Since the late 1970s, a growing number of studies provide reliable measurement of sleep quality by translating bio-information. PSG (Polysomnography) has been used as the gold standard in medicine that pastes electrodes on a patient's skin to measure brain waves, muscle tone, eye movements, breathing patterns and blood oxygen levels. However, the various sensors cause uncomfortable sleeping. Akane rapidly reduced the sensors by using the property of Electrodermal activity (EDA) that has high frequency peak patterns during deep sleep [10]. A ring with sensors measured EDA, temperature and motion data to distinguish sleep stages [11]. The motion data was used in sleep or wakefulness identification with Cole's algorithm [12]. The shortage of bio-information measurement is the need of attached sensors and is not easy used in home context.

### 2.2. Motility for sleep quality measurement

Szymansky started the first sleep monitoring system with an object method to monitor motility [13]. Many methods from different technical aspects, like electromyogram (EMG) of PSG, image sequence analysis, or pressure-sensitive mattresses, have been used to monitor activity during sleep. A historical statistical method, actigraph, was proposed by Tryon [14]. Based on the actigraph method, an actigraphy is a wearable device with accelerator and light sensors that detects physical activities to identify a sleep or wakefulness state. It samples the motion data frequently which are aggregated at a constant interval as an epoch. In PSG monitoring, an actigraphy is worn on a patient's wrist for comparing epoch-by-epoch with the PSG data. Wakemate is a commercial product with the actigraph technology that synchronizes motion data and a cellphone to make a personalized alarm by determining the lightest phase of sleep [15]. Many studies have aimed on a non-invasive technique for sleep activity monitoring. Wang designed a piezopolymer film (PVDF) sensor which was placed under a sheet at the location of thorax to monitor activities of respiratory movements and heartbeats [16]. Nakajima adapted optical flow method to visualize the velocity filed of entire body, including breast movements of respiration and posture changes. The motions consist of respiration, cessation of breath, full posture change, limb movement, and out of view [17].

There is a correlation between motions and sleep states. In an evaluation of total sleep time, a correlation of 0.81 between measurement of actigraphy and PSG in long-term period proves that an actigraphy can be used for sleep monitoring on its own [18]. In this study, we monitor entire body and consider the magnitude of body movement to be the current sleep state.

## 3. SYESTEM DESIGN

A non-invasive monitoring system is developed by an event-based method for quantitative evaluation of sleep disturbances. A device (Kinect for Windows) with a depth sensor, a RGB camera, and a microphone was used in our sleep monitoring system [18]. RGB images use 8-bit VGA resolution (640 × 480 pixels), depth signals use 11-bit VGA resolution (640 × 480 pixels) which provides 2,048 levels of sensitivity, and audio signals are at 24-bit resolution. Both of depth signals and RGB images are at a frame rate of 30 Hz, and Audio signals are with 16 kHz sampling rate. Besides, for reducing noises, a region-of-interest (ROI)

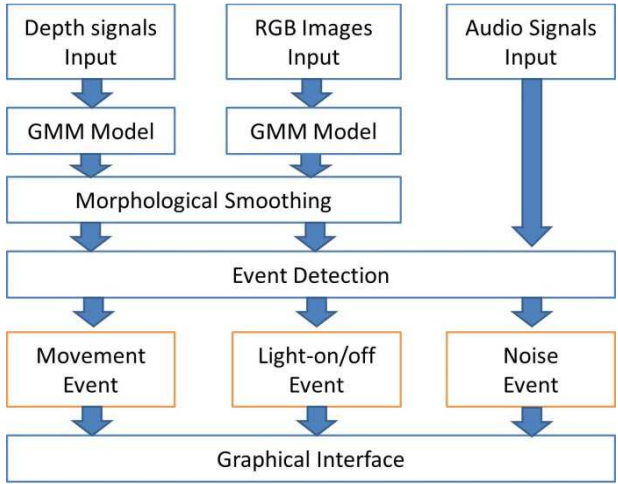

Fig. 2: System Architecture.

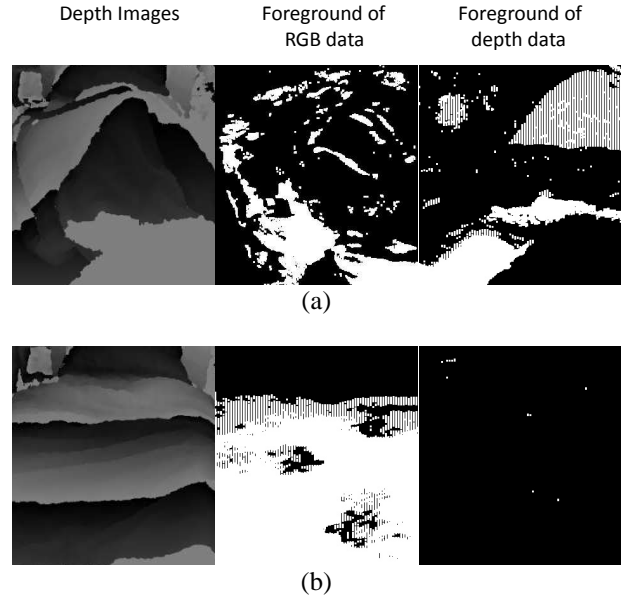

Fig. 3: Foreground Segmentation in Sleep Condition: (a) A scenario of turning behavior, (b) A scenario of turning on a light.

($320 \times 350$ pixels) is selected which covers a bed in monitoring.

The system architecture is as follows (Figure 2). Motion events, light on/off events and noise events are detected from depth signals, RGB images and audio signals respectively. One GMM background model is established in depth signals for activity detection, and another is established in RGB images for detecting lighting changes [19]. Both of the segmented foreground objects are processed by morphological smoothing to reduce noises. Then the data from three types of sensors are scored simultaneously and are normalized to the same scale. Finally, events are detected from the scores by an event detection method with empirical thresholds. According to the categories of events, the system records the captured depth images, RGB images, or waveform audio files. A graphical interface is provided with browsing functions that a specific section of graphs can be expanded and amplified for sleep examination.

### 3.1. Background models for activities and lighting changes detection

The problem of activity monitoring in a sleep environment can be regarded as foreground detection. A general approach to face it is to build a background model. In raw depth data with a problem of missing bits and flickering issues from multiple reflections, transparent objects, or scattering surfaces that the traditional background subtraction method failed for the high variety conditions. When a camera is fixed, the optical flow method is an option for sleep monitoring. However, the nature of the optical flow method with a hypothesis of a constant brightness pattern that is useless to depth signals. The concept of Gaussian Mixture Model (GMM) is to model each background pixel as a mixture of Gaussians and using sufficient statistics to update the model. The GMM method has been tested in both depth images and RGB images that the depth signals helps to establish a more stable background model than that of the RGB images, because the depth signals have the immunity against illumination changes or lack of contrast (Figure 3).

Many studies of human detection use information of depth signals or that both in depth signals and RGB images [20]. However, it is not possible to detection human shape in sleep scenario for body with quilt covering and an almost darkness room. In this work, we adopt GMM model and take the advantage of complementary data from depth signals and RGB images. In contrast with RGB images, the better performance of activity detection in depth signals and an environment with barely light source. On the contrary, RGB images are used for detecting lighting changes because depth signals can't observe the changes. After morphological smoothing, areas of foreground objects in depth signals and RGB images are scored. Besides, audio signals are scored simultaneously.

Finally, we regard the scores of depth signals and RGB signals as magnitude of movements and lighting effect respectively, and both of them and audio scores are in the same scale after normalization.

### 3.2. An epoch approach for event detection

We propose an event detection algorithm based on an epoch method for determining occurrences of events. Instead of an instant moment, the concept of epoch method aggregates data at a constant interval for recording continuous events. The period of an epoch is one second. A number of three types of scores above empirical thresholds are calculated per second. An event is determined by comparing the number of current epoch with that of the preceding epoch and the following epoch. An iterative equation is listed below, where $E_0$ is a current epoch, and $E_{-1}$ and $E_1$ are the preceding epoch and following epoch respectively.

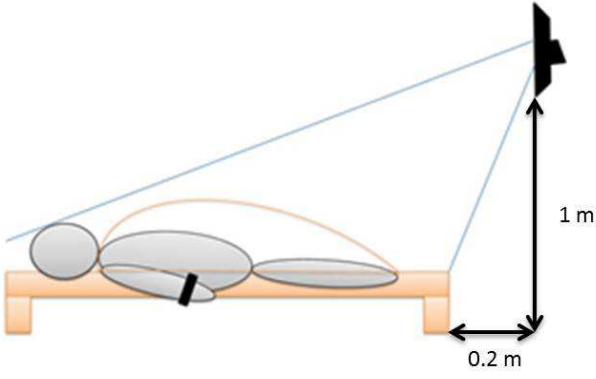

Fig. 4: Experiment Setting.

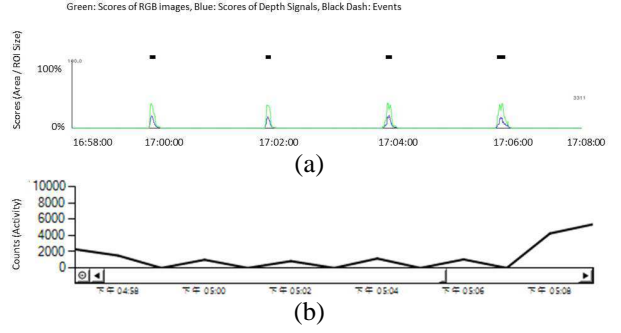

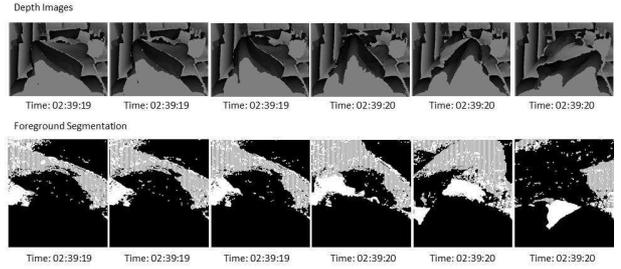

Fig. 5: A Position Change Test :( a) Sleep Monitoring System, (b) Actigraphy.

$$\begin{cases} if\ E_{-1} < E_0, start\ of\ an\ event \\ if E_0 \leq E_1, the\ event\ continues \\ ;\ otherwise, end\ of\ the\ event \end{cases} \quad (1)$$

If the number of current epoch is bigger than that of the previous epoch, an event is started of recording. The event continues when the number of the following epoch is equal or bigger than that of the current epoch, otherwise the event is ended of recording. In this work, scenarios with full posture changes, turning on/off a light, talking voices are regarded as events for recording.

## 4. EXPERIMENT

Experiments were set in home context that participants were allowed to lie on their own beds in a darkness room or with a small lamp. We think that is helpful to their sleeping and also avoids an effect of unfamiliar environment which usually causes trouble sleeping. The device (Kinect for Windows) was placed at a distance of 0.2 meters and in a height of 1 meter. It set in front of a bed at a depression angle of 27° that the camera view covers the entire body of a participant. Meanwhile, an actigraphy (ActiSleep) was worn on right hand or left hand of a participant [21]. The experiments include two parts: a posture change test in a light room and a sleeping condition in a darkness room. A 57 years old female participated in the former test and worn the actigraphy on her right hand (a left-handed person), and a 32 years old female participated in the later experiment and worn the actography on her left hand.

### 4.1. A posture change test for motion behavior measurement

For determining thresholds in motion detection, a ten minutes posture change test was designed. The participant was asked for lying on her back with the face up in a light room. An alarm was directed her to turn her body over every two minutes. After the last turning behavior, the test was ended in a two minutes rest.

For observing the movements, we only show the waveforms of depth signals and RGB images on the graph, and in this case, the waveform of RGB images is also considered as magnitude of movements in a light environment. The waveforms with frequent peaks correspond to the posture change behavior (Figure 5 (a)). Specially, the waveform of RGB images with clear peaks than that of depth signals is due to a larger noise within the RGB images stream. The corresponding image frames of turning behavior were recorded (Figure 6). The actigraphy counts the number of activities by an accelerometer sensor (Figure 5 (b)). To compare the result of our system with that of the actigraphy, the turning behavior appeals in both of the results with similar patterns. After the experiment, the thresholds for motion detection were determined.

### 4.2. A sleeping condition for sleep monitoring

For sleep monitoring, two quite different samples were captured from a participant that one is a record of trouble sleeping and another is a record of successful sleeping.

Here shows a record of trouble sleeping for one hour that the participant had trouble to fall asleep and stayed awake during the entire monitoring until she gave up sleeping (Figure 7 (a)). In sleep condition, the waveform of RGB images is considered as magnitude of lighting effects in a dark environment. The sharp waveforms on the graph are with many large peaks. The black dashes on the top of graph indicate events from three types of data. A segment with frequent peaks is expanded and amplified for examining the magnitude of movements (Figure 7 (c)). The large peaks were marked as motion events with full posture changes and the corresponding image frames were recorded. The smaller peaks can be regarded as limb movements. A flat waveform between very large peaks discovered that the participant was out of view for a while (Figure 7 (d)). A motion event recorded that the participant left the bed, then a light on/off event recorded that she turned on a light. The waveform of the actigraphy well recorded the motion behavior of the participant (Figure 7 (b)). Here shows a quantitative analysis of the trouble sleeping (Table 1). The percentages of the three types of events are listed, and we close analyze a motion event by the empirical thresholds. The components of a motion event include full posture changes, limb movements, tiny movements, calmness and out of view. We consider the full posture changes, limb movements and out of view as wakefulness, like turning behavior, scratching, or leaving a bed. The rest are identified as sleep, in which the tiny movements represents unaware movements, like small twitches in REM sleep, and the calmness means that the participant keeps a steady position in deep sleep. Sleep efficiency is the total sleep time which is divided by the total time in bed. To compare the result of our system and that of the actigraphy, the sleep efficiency is more precise by the actigraphy with Sadeh's algorithm than that with Cole's algorithm that calculates none of sleeping, because the participant did not once fall asleep. The obvious high sleep efficiency of our system is due to the different sleep or wakefulness identification between the actigraphy and our system. Even a tiny movement, like twitches in REM sleep will be identify as wakefulness by the actigraphy when the participant is sleeping. In our system, these tiny movements are regarded as regular movements in deep sleep. One situation caused the errors when the participant remained motionless without sleeping. Especially, some people if they had been trained in relaxation techniques of yoga mediation, and that is the case in the experiment. The experiment result shows that our system well detects the frequent events of trouble sleeping, but has errors with sleep efficiency in a short period monitoring.

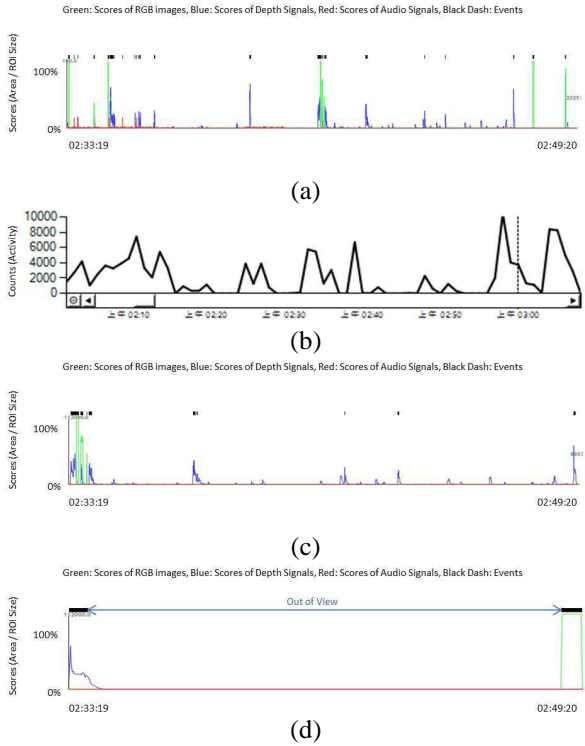

Fig. 7: A Trouble Sleeping for One Hour: (a) Sleep Monitoring System, (b) Actigraphy, (c) Expansion of Curve, (d) Out of View.

A record of the successful sleeping lasted for 6 hours with fewer events than the trouble sleeping (Figure 8 (a)). Motion events occurred frequently in the beginning of monitoring that the participant had not already sleep. After the participant falling asleep, only few motion events were detected as turning behavior. Instead of nature wake-up, the participant woke by an alarm that explains no frequent motion events in the end of monitoring. Besides, the interruptions from others were eliminated for the participant slept alone. Only one light on/off event was detected when the participant woke up and turned on a light. The sleep disturbances are obvious fewer in the experiment than that of the trouble sleeping. To compare the result of our system with that of the actigraphy, both of them recorded the motions in the beginning of sleeping, and a large peak in the end of sleeping indicates that the participant woke up and left the bed (Figure 8 (b)). Here shows a quantitative analysis of the successful sleeping with better sleep quality than that of the trouble sleeping (Table 2). The percentages of full posture changes and limb movements are less than that of the trouble sleeping. The slightly higher sleep efficiency of our system than that of the actigraphy is caused by the same reason that we regard tiny movements as regular movements in deep sleep. However, the errors with the sleep efficiency reduce in the monitoring that the sleep efficiency of our system is close to that of the actigraphy with both algorithms. The experiment result shows that our system performs well with both the events detection and evaluation of sleep efficiency in the overnight monitoring.

Table 1: Sleep Analysis of Trouble Sleeping.

| Our System | | | | | | | Sleep Efficiency | Actigraphy by Cole Sleep Efficiency | Actigraphy by Sadeh Sleep Efficiency |
|---|---|---|---|---|---|---|---|---|---|
| Motion Event (Total: 100%) | | | | | Light on/off Event | Noise Event | | | |
| Full Posture Changes | Limb Mov. | Tiny Mov. | Calmness | Out of View | | | | | |
| 1.70% | 7.10% | 46.05% | 35.13% | 10.00% | 0.93% | 0.03% | 81.2% | 8.33% | 0% |

Table 2: Sleep Analysis of Successful Sleeping.

| Our System | | | | | | | Sleep Efficiency | Actigraphy by Cole Sleep Efficiency | Actigraphy by Sadeh Sleep Efficiency |
|---|---|---|---|---|---|---|---|---|---|
| Motion Event (Total: 100%) | | | | | Light on/off Event | Noise Event | | | |
| Full Posture Changes | Limb Mov. | Tiny Mov. | Calmness | Out of View | | | | | |
| 0.26% | 3.7% | 53.26% | 40.5% | 2.25% | 0.04% | 0.15% | 93.79% | 91.6% | 89.92% |

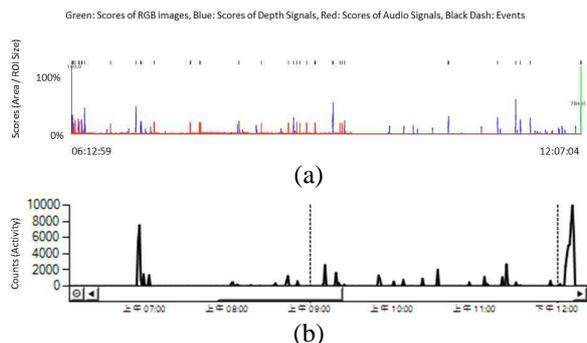

Fig. 8: A Successful Sleeping for Six Hours: (a) Sleep Monitoring System, (b) Actigraphy.

## 5. CONCLUSION AND FUTURE WORK

A non-invasive monitoring system is developed by an event-based method for quantitative evaluation of sleep disturbances. We classify the sleep disturbances into three types of events, including motion events, light on/off events and noise events. By establishing background models, magnitude of movements and lighting effects can be quantified. All of these events are detected by an event detection method. The experiment result shows our system with reliability that can evaluate sleep patterns in an overnight sleeping condition.

We'll apply the system in a clinical research that observes sleeping of elder people who after chemotherapy regiments. The collections of clinical counterparts will be helpful for us to discover need of patients with sleep disorders. We'll consider that to provide helpful guides or advices in our system for improving their sleep qualities.

## ACKNOWLEDGEMENT

This work was partially supported by the National Science Council, Taiwan, under grants NSC 101-2221-E-002-212.